\documentclass[letterpaper, 10 pt, conference]{ieeeconf}  

\IEEEoverridecommandlockouts                              

\usepackage{url}
\usepackage{amsmath}
\usepackage{algorithm}
\usepackage{algorithmic}
\usepackage{xcolor}
\usepackage{multirow}
\usepackage{graphicx}
\usepackage{subcaption}
\usepackage{hyperref}

\title{\LARGE \bf
A Unified Framework for Empowerment and Predictive Control
}

\author{Wooyoung Chung$^{1,*}$, Tristan Shah$^{1,*}$, Volodymyr Makarenko$^1$, Stas Tiomkin$^{1}$
\thanks{$^{*}$Equal contribution.}
\thanks{$^{1}$Department of Computer Science, Whitacre College of Engineering, Texas Tech University. {\tt\small \{trisshah, wochung, vmakaren, stas.tiomkin\}@ttu.edu}}
}

\begin{document}

\maketitle
\thispagestyle{empty}
\pagestyle{empty}


\begin{abstract}

Sampling-based model predictive control (MPC) is a powerful approach to trajectory optimization, but its performance depends on an informative cost function that often requires substantial domain knowledge to design. An alternative is to derive objectives directly from the system dynamics. Empowerment, defined as the channel capacity between an agent’s inputs and its future state, provides such a goal-agnostic objective and has been shown to produce useful behaviors across a range of domains. However, existing empowerment-based controllers have two key limitations: empowerment is estimated using a virtual probing policy distinct from the executed control policy, obscuring its connection to the resulting behavior; and its computation typically requires second-order derivatives of the dynamics, limiting compatibility with standard MPC methods. We formulate empowerment using a single policy for both probing and acting, directly linking the empowerment objective to the executed behavior. The resulting objective requires only first-order derivatives and can be optimized with standard MPC solvers, either alone or in combination with a task-specific cost. We evaluate the proposed approach on classical control tasks and show that combining empowerment with task costs matches or exceeds the success rate achieved by either objective alone. Our formulation provides a practical bridge between standard MPC and intrinsically motivated control.

\end{abstract}


\section{Introduction}
\label{sec:intro}

Model Predictive Control (MPC) \cite{richalet1978mpc} is a widely used control paradigm in which future system behavior is predicted using a dynamical model and optimized over a finite time horizon. At each control step, MPC solves a finite-horizon optimization problem to predict outcomes, enforce constraints, and minimize a specified cost function, while executing only the first action before re-planning. Owing to its flexibility and principled handling of constraints, MPC is a powerful and practical tool for robotic control, particularly for complex, nonlinear, and high-dimensional systems. In this work, we consider sampling-based MPC, which evaluates candidate action trajectories through forward simulation and performs derivative-free optimization over the control sequence. Its performance, however, depends on an informative cost function. A cost that encodes only the task objective may be sparse, incomplete, or non-informative over large regions of the state space. It is therefore common to introduce additional terms, such as penalties on control effort, state deviation, or instability, to shape the optimization problem~\cite{ibrahim_2024}. These terms are typically task-specific and hand-tuned, and their design can require substantial domain knowledge.

\begin{figure}
\centering

\begin{subfigure}{\linewidth}
    \centering
    \includegraphics[width=\linewidth]{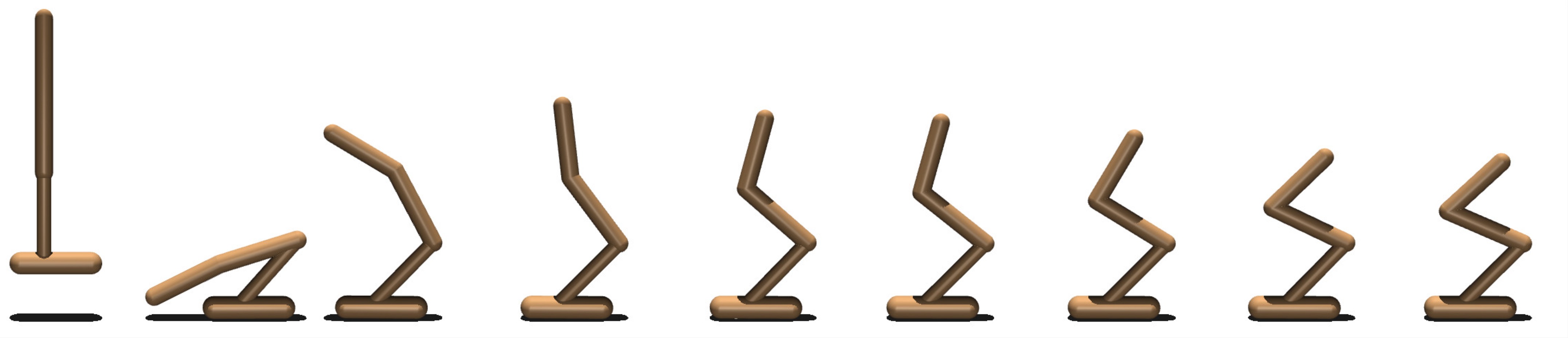}
    \caption{Self-recovery from a fallen configuration using MPC-based empowerment.}
    \label{fig:side_by_side}
\end{subfigure}
\vspace{0.5cm}
\begin{subfigure}{\linewidth}
    \centering
    \includegraphics[width=\linewidth]{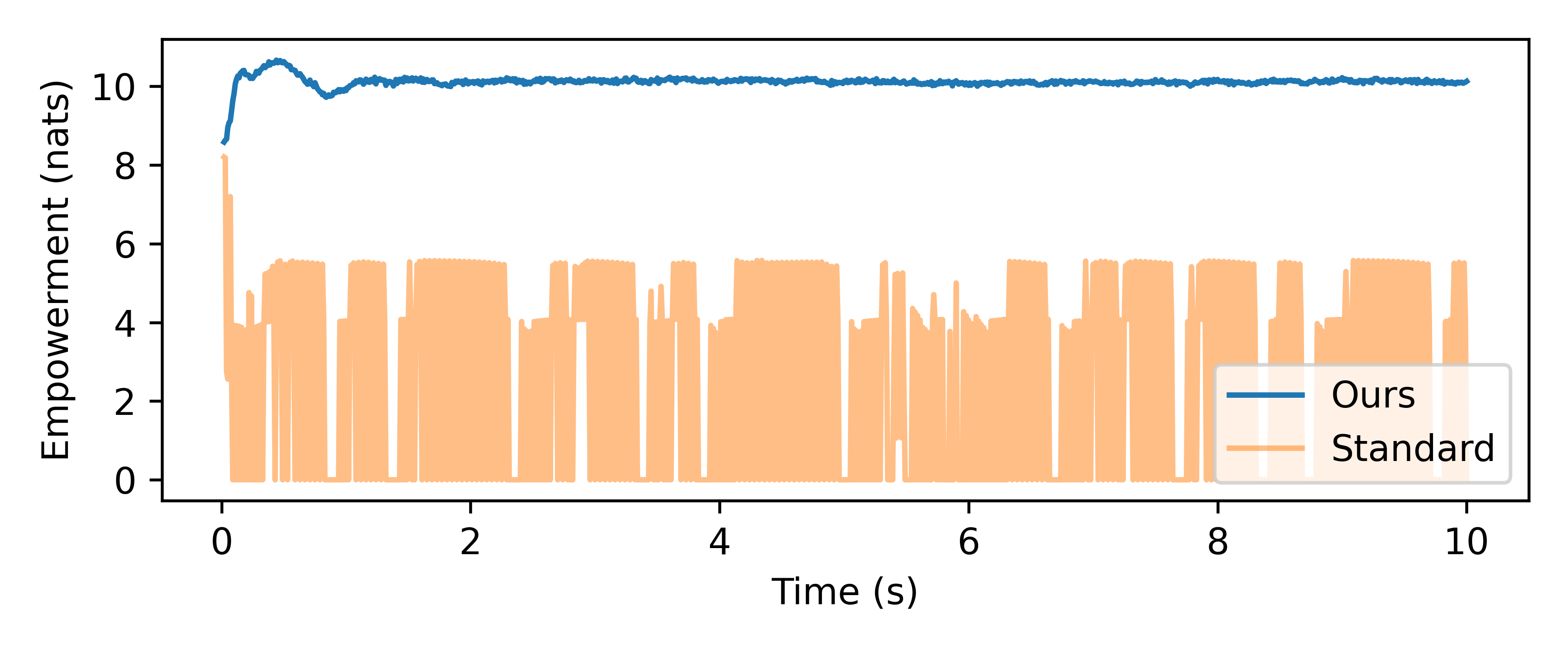}
    \caption{MPC-based empowerment remains stable during self-recovery, unlike standard empowerment.}
    \label{fig:empowerment_values}
\end{subfigure}
\vspace{-0.85cm}
\caption{{\bf Goal-agnostic self-recovery using MPC-based empowerment.} 
(a) The proposed MPC-based empowerment controller recovers from a fallen configuration and stabilizes the hopper without a task-specific recovery objective. 
(b) The corresponding empowerment values remain stable under MPC-based empowerment, while standard empowerment, which relies on second-order derivatives for control, exhibits large oscillations and fails to stabilize the system.}
\label{fig:hopper_stabilization}

\end{figure}

An alternative is to derive the optimization objective directly from the system dynamics. Empowerment is one such objective, formalized as the channel capacity between an agent’s actions and its future states \cite{salge2014empowerment, tiomkin2024intrinsic}. This information-theoretic objective has been shown to induce meaningful behaviors across a range of settings, including discrete \cite{Tiomkin_2025} and continuous control \cite{karl2017unsupervisedrealtimecontrolvariational, tiomkin2024intrinsic}, embodied agents, and multi-agent systems \cite{kopio2023improving}. These results motivate empowerment as a general-purpose, goal-agnostic control objective. Existing empowerment-based control methods, however, have important practical limitations. They typically rely on two distinct policies or action distributions: one for ``probing’’ the dynamics to estimate empowerment and another for executing control. This separation creates a mismatch with sampling-based MPC, which evaluates and selects control actions using a common trajectory distribution. It can also increase sampling requirements and obscure the relationship between the empowerment objective and the resulting control behavior.

We address this gap by formulating empowerment as a trajectory-level objective within sampling-based MPC. Our approach unifies probing and control into a single action distribution, allowing the same sampled trajectories to support empowerment estimation and control optimization. This improves sampling efficiency and directly connects the empowerment objective to the executed control behavior. 

Importantly, sampling-based MPC performs derivative-free optimization over the control sequence. Derivatives are required only to estimate empowerment through sensitivity analysis of the dynamics. Standard empowerment-based controllers require an additional differentiation of the empowerment estimate to obtain the control law, resulting in second-order derivatives of the dynamics \cite{tiomkin2024intrinsic}. By instead optimizing empowerment directly as an MPC objective, our formulation requires only the first-order sensitivities used to estimate empowerment. This yields a natural integration of empowerment with derivative-free trajectory optimization while reducing the derivative order required for control.

Our main contributions are as follows:
\begin{enumerate}
\item We introduce an empowerment formulation based on a \emph{single} action distribution for both probing and control, directly connecting empowerment estimation to the executed control behavior.
\item We integrate empowerment with sampling-based MPC, using first-order dynamics sensitivities to estimate empowerment and derivative-free trajectory optimization for control. This avoids the second-order derivatives required by existing empowerment-based controllers \cite{tiomkin2024intrinsic} and improves sampling efficiency by using the same trajectory samples for empowerment estimation and control.
\item We show that the proposed formulation naturally combines intrinsic empowerment with task-specific costs along the same trajectories, and demonstrate that the combined objective can improve control performance relative to either objective alone.
\end{enumerate}

The remainder of the paper is organized as follows. Sections \ref{sec:emp} and \ref{sec:mpc} review empowerment-based control and sampling-based predictive control, respectively, and identify the algorithmic gap between the two formulations. Section \ref{sec:method} presents our MPC-based empowerment formulation and its integration with task-specific control objectives. Section \ref{sec:experiments} provides an empirical evaluation of the proposed method. Finally, Section \ref{sec:discussion} discusses the results, limitations, and directions for future work. Our code is publicly available on \href{https://github.com/tristan-shah/empowerment-predictive-sampling}{\color{magenta}\nolinkurl{GitHub}}.

\section{Empowerment}\label{sec:emp}

Standard empowerment-based controllers \cite{1554676, salge2014empowerment,leibfried2019bellman,massari2021empowerment,dai2020empowerment,cao2025causal,tiomkin2024intrinsic,volpi2016computation} rely on two separate action distributions, one for evaluating empowerment by probing, and the other to derive optimal actions. In this section, the linear Gaussian formulation for empowerment estimation and control \cite{tiomkin2024intrinsic} is shown in order to highlight this discrepancy. Our proposed solution, which unifies probing and acting, is introduced in Section~\ref{sec:method}.

Given nonlinear discrete-time dynamics:
\begin{equation}\label{eq:dynamics}
    x_{t+1} = f(x_t, u_t),
\end{equation}
with states $x_t\in X^{d_x}$ and actions $u_t\in U^{d_u}$. Empowerment of the state, $x_0$, is defined as the maximal mutual information between future states and action sequences:
\begin{align}
    \mathcal{E}(x_0) = \max_{\pi(\cdot|x_0)} \mathcal{I}[X_T; U_{0:T-1}\mid x_0]. \label{eq:emp def}
\end{align}  
We use capital and small letters for random variables and their specific realizations, respectively. For example, $X$ and $U$ are RV, while $x$ and $u$ are their respective specific values. Here: 

In Equation \eqref{eq:emp def}, maximization is over the probing policy $\pi(U_0^{T-1} | x_0)$ which explores virtual open-loop action sequences where $U_{0:T-1} = \begin{bmatrix} U_0, \cdots, U_{T-1} \end{bmatrix}$. The result of the channel capacity calculation in Equation \eqref{eq:emp def} is interpreted as the maximal achievable volume of final states $x_T$ by the optimal choice in probing distribution. Effectively, this represents the number of options available to the agent in the future.


There are three steps in the derivation of empowerment based controllers in continuous domains, shown in the three following subsections and summarized in the Algorithm \eqref{alg:two policies}. 
\subsection{Dynamics as a Communication Channel}
A state $T$-steps in the future, $x_T$, driven by a control sequence, 
can be expressed by iterated dynamics, 
\begin{equation}
    x_{T} = f^{T-1}(x_0, u_{0:T-1}),
\end{equation}
where $f^{T-1}$ is the $T-1$ times iterated dynamics in Equation \eqref{eq:dynamics}. The final state $x_T$ is expressed as a linear Gaussian channel through linearization of $f^{T-1}$ around zero actions, $\mathbf{0}\doteq\mathbf{0}_{0:T-1}$: 
\begin{align*}
    x_{T} &= f^{T-1}(x_0, u_{0:T-1}) \\
    & \approx f^{T-1}(x_0, \mathbf{0}) + F_T(x_0, \mathbf{0})\cdot u_{0:T-1},
\end{align*}
where $F_T(x_0, \mathbf{0})$ is the Jacobian of $f^{T-1}$ with regard to $u_{0:T-1}$, evaluated along a nominal action sequence $\bar{u}_{0:T-1}=\mathbf{0}$ in state $x_0$. To ensure finite channel capacity a small amount of noise is added to the final state, $\epsilon \sim \mathcal{N}(\mathbf{0}, \Sigma_{\epsilon})$:
\begin{align}
    x_T \approx f^{T-1}(x_0, \mathbf{0}) + F_t(x_0, \mathbf{0}) u_{0:T-1} + \epsilon \hfill&&\text{(\textbf{Channel})}\label{eq:channel}
\end{align}
The probing policy $\pi$ is assumed to be a multivariate Gaussian distribution centered on a mean control sequence of zero actions:
\begin{equation}\label{eq:zero_mean_gaussian_probing}
    \pi(U_{0:T-1} | x_0) := \mathcal{N}(\mathbf{0}, \Sigma_{u_{0:T-1}}),
\end{equation}
where $\Sigma_{u_{0:T-1}}$ is the covariance matrix of the open-loop control sequence.

\subsection{Channel Capacity via Water-filling}
Applying Equation \eqref{eq:channel} to Equation \eqref{eq:emp def} under the assumption of a Gaussian probing policy in Equation \eqref{eq:zero_mean_gaussian_probing} enables the calculation of empowerment by the water-filling algorithm. In this setting, the  optimization is over the covariance matrix of the probing policy, $\Sigma_{u_{0:T-1}}$:
\begin{equation}\label{eq:emp LGC}
\begin{aligned}
\mathcal{E}(x_0, \mathbf{0})
&= \max_{\Sigma_{u_{0:T-1}}}
\frac{1}{2}\ln \lvert \Sigma_{x_T} \rvert, \quad\text{(\textbf{Water-filling})}\\
\Sigma_{x_T}
&= F_T(x_0, \mathbf{0})
   \Sigma_{u_{0:T-1}}
   F_T(x_0, \mathbf{0})^\top
   + \Sigma_{\epsilon}
\end{aligned}
\end{equation}
where $\Sigma_{x_T}$ is the covariance of the final state $x_T$. For notational simplicity we omit the functional dependency of $\Sigma_{x_T}$ on $x_0$ and $\bar{u}_{0:T-1}$ when clear from context. The solution in Equation \eqref{eq:emp LGC} returns the value of empowerment (channel capacity) under the optimal probing covariance matrix \cite{cover1999elements}.

Crucially, in this formulation the water-filling algorithm optimizes only $\Sigma_{u_{0:T-1}}$, while the mean action sequence, $\bar{u}_{0:T-1} = \mathbf{0}$, remains unaffected. In our novel method, the latter restriction is removed to allow $\bar{u}_{0:T-1}$ to affect channel capacity.
    
\subsection{Controller via The Empowerment Principle}

The empowerment principle posits that meaningful behavior emerges when an agent pursues states with high empowerment \cite{1554676,salge2014empowerment,tiomkin2024intrinsic}. This principle suggests a controller based on gradient ascent:
\begin{align}
    u_t^*(x) \propto f_{u}^\top \cdot \nabla_x\mathcal{E}(x, \mathbf{0})\hfill&&\text{\textbf{(Action)}}\label{eq:opt action}
\end{align}
where the control gain $f_{u} = \frac{\partial f(x, u)}{\partial u}$ is the partial derivative of the dynamics with respect to control. It provides a heuristic on how to map the gradient of empowerment with respect to the state into the control space of the agent.

The optimal action in Equation \eqref{eq:opt action} is applied to the dynamics, which increases empowerment, $x_{t+1}=f(x_t,  u_t^*(x))$. These three steps, (\textbf{Channel, Water-Filling, Action}) are repeated at $x_{t+1}$. This scheme (summarized in Algorithm~\ref{alg:two policies}) has been demonstrated in a range of control problems \cite{1554676, salge2014empowerment,massari2021empowerment,cao2025causal,tiomkin2024intrinsic}. 
\begin{algorithm}[t!]
  \caption{Classical Empowerment Controller}
  \label{alg:two policies}
  \begin{algorithmic}[1]
    \STATE {\bfseries Input:} model $f$, current state $x$, planning horizon $T$
    \REPEAT
    \STATE Estimate empowerment as channel capacity:
    \STATE $\;\;\;\mathcal{E}(x, \mathbf{0}) \gets \max_\pi\mathcal{I}[X_{T}; U_{0:T-1}| x]$
    \STATE Derive action in the direction of the gradient:  
    \STATE $\;\;\;u^*(x) \propto f_{u}^\top \cdot \nabla_x \mathcal{E}(x, \mathbf{0})\hfill \text{($2^{nd}$ order controller)}$
    \STATE Apply the action to dynamics:
    \STATE $\;\;\; x \leftarrow f(x, u^*(x))$
    \UNTIL{convergence at salient states}
  \end{algorithmic}
\end{algorithm}

One limitation of this formulation is that the actual control action, $u_t^*(x)$, in Equation \eqref{eq:opt action}, and the virtual actions from the probing policy in Equation \eqref{eq:zero_mean_gaussian_probing} are disconnected. The virtual actions do not drive dynamics, but rather, they probe the channel around the nominal trajectory induced by the constant zero action sequence. In contrast, in predictive control, the cost is evaluated under actual control sequences, which are used in the dynamics.

\section{Model Predictive Control}\label{sec:mpc}

Sampling-based (zero-order) predictive control methods \cite{williams2017mppi, howell2022predictive, pinneri2021sample} optimize finite-horizon action sequences by directly evaluating candidate rollouts under a forward model, avoiding reliance on gradients of the dynamics \cite{richalet1978mpc, falcone2007mpc, hamalainen2015pbp, audet2017derivativefree, na2003mpc}. 
These planners maintain a distribution over actions, sample trajectory rollouts, and update the distribution to maximize predicted cumulative return. This makes them particularly effective for nonlinear, discontinuous, or contact-rich systems, while their natural amenability to parallel simulation renders them well suited for real-time MPC.


A key challenge in combining MPC with existing empowerment methods is the mismatch in how action distributions are used. Standard empowerment formulations \cite{tiomkin2024intrinsic, karl2017unsupervisedrealtimecontrolvariational} estimate channel capacity through virtual probing actions, while control actions are derived separately, typically through empowerment gradients. This separation leads to incompatible action representations, increased sampling complexity, and difficulties integrating empowerment with MPC objectives that rely on a single action distribution to jointly evaluate future costs and select actions. Bridging this gap requires reformulating empowerment so that both empowerment and the task objectives can be evaluated coherently within the same predictive sampling framework.






\section{Proposed Method}\label{sec:method}

In classical empowerment \cite{tiomkin2024intrinsic}, the nonlinear system dynamics are linearized around a zero control sequence to form a tractable communication channel. The resulting nominal trajectory is passive, while a probabilistic probing policy $\pi$ explores action sequences within a local neighborhood (or ``tube'') around this trajectory. This assumption is reasonable for sufficiently small control magnitudes, where the linear approximation remains accurate. However, this formulation introduces ambiguity in selecting the actions actually executed by the agent, typically necessitating heuristic action selection rules such as those in Equation \eqref{eq:opt action}. Moreover, classical empowerment does not naturally support planning over trajectories or straightforward integration with task objectives.

In our framework, classical empowerment is recast as a trajectory optimization problem by allowing the probing policy to have a nonzero mean. Specifically, the probing policy in Equation \eqref{eq:non_zero_mean_probing_policy} is defined as a Gaussian distribution centered at a mean control sequence $\bar{u}_{0:T-1}$, which need not be zero. This formulation enables the probing policy to explore locally around a nominal trajectory induced by $\bar{u}_{0:T-1}$, while simultaneously allowing the mean control sequence itself to be optimized so as to guide the agent toward states with higher empowerment.
\begin{equation}\label{eq:non_zero_mean_probing_policy}
    \pi(U_{0:T-1} | x_0) := \mathcal{N}(\bar{u}_{0:t-1}, \Sigma_{u_{0:t-1}})
\end{equation}
The mean control sequence $\bar{u}_{0:T-1}$ is iteratively updated to maximize Equation \eqref{eq:nonzero_mean_empowerment}, and the first action of the optimized sequence is applied to the system.
\begin{equation}
    \mathcal{E}(x_0, \bar{u}_{0:T-1}) = \max_{\Sigma_{u_{0:T-1}}}\frac{1}{2}\ln\big|\Sigma_{x_T}\big|\label{eq:nonzero_mean_empowerment}
\end{equation}
The covariance matrix of the final state $\Sigma_{x_T}$ is given by Equation \eqref{eq:state_covariance}, which demonstrates the functional dependency on the mean control sequence.
\begin{equation}\label{eq:state_covariance}
    \Sigma_{x_T} = F_T(x_0, \bar{u}_{0:T-1})\Sigma_{u_{0:T-1}}F_T(x_0, \bar{u}_{0:T-1})^\top + \Sigma_{\epsilon}
\end{equation}

\subsection{Novel Formulation for Single-Policy Empowerment}
We formulate empowerment estimation as the following nested optimization problem: 
\begin{equation}
    \mathcal{E}(x_0) = \max_{\Sigma_{u_{0:T-1}}} {\color{blue}\max_{\bar{u}_{0:T-1}}} \frac{1}{2}\ln |\Sigma_{x_T}|,\label{eq:nexte emp}
\end{equation}
where the inner optimization shown in blue is over the mean of the action distribution $\bar{u}_{0:T-1}$, which affects the channel, $F_T$, while the outer optimization is over the action distributions covariance matrices. The objective in Equation \eqref{eq:nexte emp} is a generalization of standard empowerment as it reduces to Equation \eqref{eq:emp LGC} when $\bar{u}_{0:T-1} = \mathbf{0}$.

Our proposed formulation allows the agent to affect channel capacity by the action distribution mean through modulating the channel, which does not affect mutual information otherwise. This formulation eliminates the need to consider two different policies for developing empowerment-based controllers, enabling seamless integration with predictive sampling approach as summarized by Algorithm \eqref{alg:mpc-empowerment}.

As in Equation \eqref{eq:state_covariance}, $\Sigma_{x_T} = \Sigma_{x_T}\bigl(x_0, \bar{u}_{0:T-1} \bigr)$ denotes the covariance of the final state. For notational simplicity, its dependence on $x_0$ and $\bar{u}_{0:T-1}$ is suppressed in the sequel. The outer maximization over $\Sigma_{u_{0:T-1}}$ is carried out using the water-filling algorithm, as in classical empowerment.

In our method, empowerment serves as an task-agnostic objective that evaluates the controllability of trajectories generated by candidate action sequences $\bar{u}_{0:T-1}$. We employ zero-order MPC techniques, specifically, Random Shooting (RS), Predictive Sampling (PS) \cite{howell2022predictive} the improved Cross-Entropy Method (CEM) \cite{pinneri2021sample}, and Model Predictive Path Integral Control (MPPI) \cite{williams2017mppi}, to optimize $\bar{u}_{0:T-1}$ with respect to this objective. Our approach unifies the probing policy with the actions executed by the agent and avoids the need for second-order dynamics derivatives. Trajectory optimization is instead performed using zero-order methods, while empowerment evaluation relies only on first-order linearizations.

The following subsection describes the algorithm used to apply empowerment as an objective within online trajectory optimization.

\subsection{Algorithm}

We formulate empowerment as an  task-agnostic objective that can be evaluated given the agent’s current state and a candidate control sequence, making it directly applicable within MPC. The resulting online MPC-based empowerment controller is summarized in Algorithm \ref{alg:mpc-empowerment}. The procedure begins with a nominal control sequence $\bar{u}_{0:T-1}$, which is iteratively refined through the agent's online interaction with the environment. At each timestep the total cost including $\mathcal{E}(x, \bar{u}_{0:T-1})$ and $C(x, \bar{u}_{0:T-1})$ is calculated on $N$ perturbed control sequences.

At each timestep, the first control input of the optimized sequence is applied, after which the sequence is shifted forward to warm-start optimization at the next state.
\begin{algorithm}[t!]
  \caption{MPC-based Empowerment Controller}
  \label{alg:mpc-empowerment}
  \begin{algorithmic}[1]
    \STATE \textbf{Input:} dynamics $f$, current state $x$, horizon $T$,
    nominal controls $\bar{u}_{0:T-1}$, number of samples $N$, task objective $C$, and intrinsic weighting $\beta$
    \REPEAT
        \FOR{$i = 1$ to $N$}
            \STATE Sample control sequence around nominal:
            \STATE $\;\;\;\bar{u}^{(i)}_{0:T-1} \leftarrow \mathrm{Sample}(\bar{u}_{0:T-1})$
            \STATE Compute combined objective for sample $i$:
            \STATE $\;\;\;J_i \leftarrow \beta \mathcal{E}(x, \bar{u}^{(i)}_{0:T-1}) + C(x, \bar{u}_{0:T-1})$
        \ENDFOR
        \STATE Update $\bar{u}_{0:T-1}$ using $\{J_i\}_{i=1}^N$
        \STATE Apply the first control input:
        \STATE $\;\;\;x \leftarrow f(x, \bar{u}_0)$
        \STATE Shift $\bar{u}_{0:T-1}$ forward by one time step
    \UNTIL{task completion}
  \end{algorithmic}
\end{algorithm}
Algorithm \ref{alg:mpc-empowerment} combines empowerment $\mathcal{E}$ and goal-related $C$ objectives. The parameter $\beta$ controls the relative importance of the task objective during optimization. When running our formulation solely with empowerment as the objective, we set $\beta = 1$ and remove the task objective $C$, yielding a purely intrinsic controller.

In the next section, we present experimental results demonstrating the application of Algorithm \ref{alg:mpc-empowerment} to control tasks in the presence and absence of an task objective.

\section{Experiments}\label{sec:experiments}
We evaluate MPC-based empowerment in two key aspects. First, we demonstrate its capability to self-recover after a fall in the hopper environment without any task-specific cost. Second, we show that empowerment enhances control performance, particularly when the performance objective is sparse and uninformative. We validate this across pendulum stabilization tasks, including the single pendulum, cart-pole, and double pendulum.

\subsection{Intrinsic Self-Recovery}

The ability of MPC-empowerment to recover its balance after being dropped from a height above the ground is shown in Figure \ref{fig:side_by_side}. Initially the hopper falls to the ground in an uncontrollable position. However, MPC-empowerment guides the agent to occupy the most controllable state. The agent is able to self-recover from the fall and move into a crouched position. The agent is able to maintain stability indefinitely in this highly controllable position. Surprisingly, empowerment discovers that the crouching is more controllable than the fully straight upright position.

While the proposed MPC-based empowerment is able to synthesize intelligent behavior, standard empowerment is unable to do so. Despite having access to second-order gradients of the dynamics, standard empowerment does not scale well to high-dimensional environments such as the Hopper. The poor performance of the hopper is demonstrated in Figure \ref{fig:empowerment_values}, which plots the corresponding values of empowerment during online interaction. MPC-based empowerment is smooth and stable, while standard empowerment is highly noisy and falls to a low value.

\subsection{Improved Uptake of Task Objective}

\begin{figure}[t!]
    \centering
    \includegraphics[width=\linewidth]{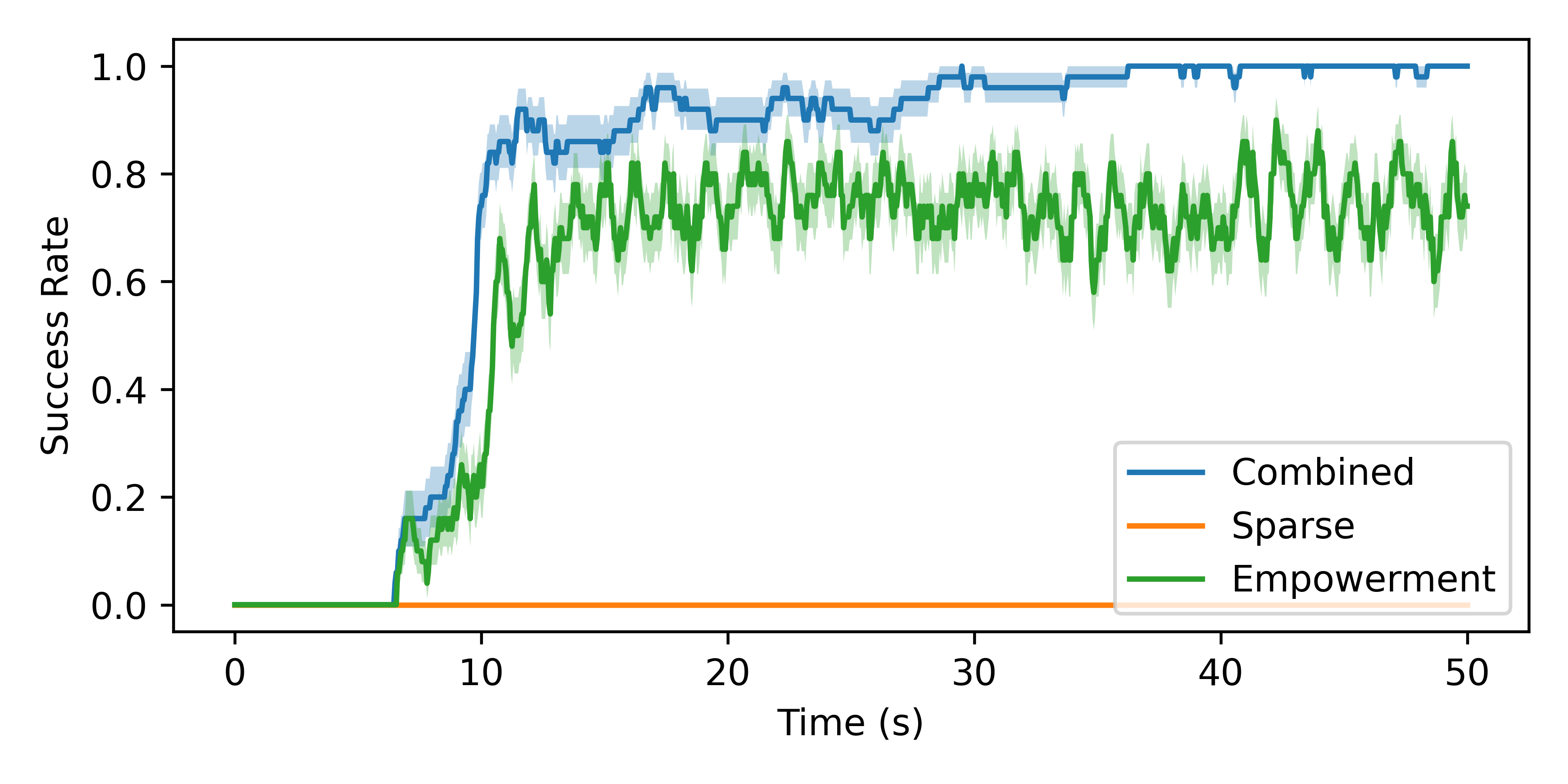}
    \caption{Success rate on cart pole during online MPC interaction with the CEM algorithm. Mean and standard error are reported over $10$ random seeds. Success rate is determined by the agent's ability to stay within a $\pi \pm \tau$ (rad) cone in the upright position. Guided strictly by the sparse task cost ($\beta = 0$), the controller fails to execute the swing-up maneuver. This is the typical behavior we observed in our experiments across all environments.}
    \label{fig:success_rate}
\end{figure}

We validate the claim that empowerment can assist in the acquisition of task objectives in Table \ref{tab:intrinsic_improvement}. In this setting we deal with \textit{sparse task costs} which are far less informative than dense ones. Sparse performance metrics are notoriously difficult to optimize due to the low probability of encountering useful error signals during state-space search \cite{Schwarke_2023}.
In both single pendulum and cart pole we define the sparse indicator cost:
\begin{equation}\label{eq:sparse_reward}
r(x_t) = \mathbf{1}
    \left\{
    \lvert \mathrm{angle}(x_t) - \pi \rvert \le \tau
    \right\}
\end{equation}
which indicates whether the pendulum is within a cone around the upright position. This objective is particularly difficult because the pendulums start in the downward position and must execute a multi-step swing-up maneuver before receiving any informative feedback signal. In all experiments we used $\tau = 0.1$. In the double pendulum, a similar objective is used, except the requirement is that both links must be in the upright position simultaneously. The overall task objective is evaluated as the accumulated sparse cost over the prediction horizon:
\begin{equation*}
    C(x_0, \bar{u}_{0:T-1}) = \sum_{t=0}^{T-1}r(x_t).
\end{equation*}
Since the instantaneous sparse cost takes takes binary values in $\{0, 1\}$, its temporal average directly corresponds to a task success rate. These are the values plotted in Table \ref{tab:intrinsic_improvement}. In Figure \ref{fig:success_rate} we plot $r(x_t)$ over time averaged over $10$ runs.

When guided strictly by the sparse task cost (setting $\beta = 0$), the MPC framework fails to swing up and stabilize any of the pendulum systems, regardless of the sampling algorithm used. As shown in Figure \ref{fig:success_rate}, optimizing solely for the task cost yields no meaningful control behavior. In contrast, MPC-empowerment alone achieves a moderately high success rate without explicit access to the task cost. This behavior is expected from empowerment controllers because they are known to seek out states that are highly controllable, such as the upright position \cite{tiomkin2024intrinsic, salge2014empowerment}, which is congruent with our selected goal state. Finally, combining empowerment with the sparse task cost improves task success rates across all sampling algorithms, as detailed in Table \ref{tab:intrinsic_improvement}.

\begin{table*}
\begin{center}

\resizebox{\textwidth}{!}{
\begin{tabular}{|c||cc||cc||cc||cc|}
\hline
\multirow{2}{*}{Task} & \multicolumn{2}{c||}{Random Shooting} & \multicolumn{2}{c||}{Predictive Sampling} & \multicolumn{2}{c||}{CEM} & \multicolumn{2}{c|}{MPPI} \\
\cline{2-9}
 & Empowerment & Combined & Empowerment & Combined & Empowerment & Combined & Empowerment & Combined \\
\hline
Single Pendulum 
& $50 \%$ & $95 \%$
& $49 \%$ & $78 \%$
& $59 \%$ & $\mathbf{100 \%}$
& $23 \%$ & $93 \%$ \\
\hline
Cart Pole 
& $22 \%$ & $79 \%$
& $81 \%$ & $\mathbf{100 \%}$
& $76 \%$ & $99 \%$
& $24 \%$ & $60 \%$ \\
\hline
Double Pendulum 
& $0 \%$ & $0\%$
& $0 \%$ & $\mathbf{86 \%}$
& $3 \%$ & $46 \%$
& $0\%$ & $0\%$ \\
\hline
\end{tabular}
}
\end{center}

\caption{
Success rate comparison between trajectory optimization guided strictly by task-agnostic empowerment ($\text{Empowerment}$) versus combined optimization with empowerment and the sparse task cost ($\text{Combined}, \beta = 0.0001$) across the single pendulum, cart-pole, and double pendulum environments. All systems are initialized in the inverted (downward) equilibrium position. The success rate is defined as the temporal average of the indicator cost in Equation \eqref{eq:sparse_reward} evaluated after the initial swing-up transient phase, with statistics computed over $10$ independent random seeds. A higher success rate reflects increased residence time within the target tolerance region around the upright equilibrium. When optimizing strictly under the sparse task cost ($\beta = 0$), all sampling algorithms fail to solve the tasks due to the low probability of sampling a swing-up trajectory that hits the target set. Notably, the combined cost formulation consistently outperforms empowerment alone across all environments.
}
\label{tab:intrinsic_improvement}

\end{table*}

\section{Discussion}\label{sec:discussion}

This work proposes a unified framework that bridges empowerment with MPC. By representing probing and control through a single action distribution, we eliminate a long-standing conceptual and algorithmic separation in empowerment-based control. This unification enables empowerment to be integrated as a primary objective within predictive control, allowing task-agnostic capacity metrics and task-specific costs to be evaluated coherently over a shared horizon. Our empirical results demonstrate that this formulation significantly improves closed-loop performance and stabilization under sparse feedback. Furthermore, it unlocks emergent control behaviors in underactuated and high-dimensional systems that were previously unexplored using empowerment.

\subsection{Unification of Probing and Control Distributions}

A central contribution of this work is the removal of the distinction between \emph{virtual probing actions} and \emph{executed control actions} that has characterized prior empowerment-based controllers. In classical formulations \cite{tiomkin2024intrinsic}, empowerment is estimated using a zero-mean probing distribution that does not reflect the system's nominal trajectory, while control actions are derived separately via gradient-based or auxiliary updates. This separation increases algorithmic complexity and fundamentally conflicts with predictive control methods, which rely on a single distribution to evaluate future outcomes and select actions.

Our formulation resolves this mismatch by allowing the probing distribution to have a nonzero mean and by directly optimizing that mean using MPC-style trajectory optimization. As a result, empowerment estimation and action execution become two aspects of the same optimization process rather than separate stages. This unification ensures that the exploratory actions used to compute channel capacity align with the control inputs applied to the system, enabling seamless integration with sampling-based MPC methods such as Predictive Sampling and the Cross-Entropy Method (CEM). Conceptually, this reframes empowerment from a local, state-dependent quantity into a trajectory-level objective that can be optimized online.




\subsection{First Order Control}

Another important advantage of the proposed framework is its computational efficiency compared to classical empowerment controllers. Previous empowerment-based controllers in continuous domains typically require second-order derivatives of the dynamics in order to map empowerment gradients into control actions. Such methods are often impractical for real-time control, especially in high-dimensional systems where second-order derivatives are expensive, numerically unstable, or unavailable.

In contrast, our approach relies on zero-order trajectory optimization for control and only first-order linearizations for empowerment evaluation. This significantly reduces computational overhead and aligns well with modern MPC pipelines used in robotics. The resulting controller is therefore better suited for real-time applications, where low latency and robustness are critical. This distinction is particularly relevant for scaling empowerment-based control to complex robotic platforms with large state and action spaces.

\subsection{Improved Uptake of Task Objective}

Our empirical results support the hypothesis that task-agnostic objectives facilitate the acquisition of task-specific performance goals \cite{schmidhuber2010formal}. Across all evaluation environments, combining empowerment with a sparse task cost consistently improves closed-loop performance, accelerating convergence to the upright equilibrium and enhancing stability compared to empowerment alone. On its own, the sparse task cost provides an insufficient gradient signal to guide optimization. Only when augmented with empowerment does the controller reliably stabilize the system.

This finding carries important practical implications for control design. Constructing dense stage-cost functions that fully specify complex system behavior is often difficult and task-dependent. Empowerment provides an objective, task-agnostic signal that naturally biases trajectory optimization toward highly controllable and stable regions of the state space, effectively smoothing the optimization landscape for model predictive control. When paired with sparse or uninformative primary costs, this auxiliary capacity metric guides trajectory sampling and improves robustness without requiring labor-intensive cost engineering.

\section{Conclusion}

In this work, we introduced a novel formulation of empowerment maximization within the framework of MPC. Our results demonstrate that empowerment can serve as an effective task-agnostic objective, enabling the synthesis of complex behaviors—such as self-righting in underactuated, high-dimensional systems—and significantly accelerating convergence under sparse performance costs. These findings indicate that empowerment-guided predictive control can induce coherent, goal-directed behavior even in the absence of dense task supervision.


Despite these promising results, several limitations remain. First, we observed that MPC-based empowerment is highly sensitive to hyperparameter choices, particularly the planning horizon. Determining an appropriate horizon remains an open challenge not only in our setting but in empowerment-based control more broadly \cite{tiomkin2024intrinsic, salge2014empowerment, karl2017unsupervisedrealtimecontrolvariational}. It is generally unclear how far into the future an agent must plan to produce stable and intelligent behavior. Importantly, increasing the planning horizon does not monotonically improve performance. This non-monotonic relationship complicates both tuning and theoretical understanding.

Second, although the proposed formulation benefits from relying on first-order rather than second-order derivatives, we found that MPC-based empowerment requires a large number of parallel rollouts to achieve stable optimization, particularly in more complex environments. In contrast, zero-order MPC guided by a well-engineered task objective can often solve challenging tasks such as locomotion and dexterous object manipulation with comparatively fewer samples. This discrepancy suggests that, while empowerment highlights controllable aspects of the state space, its optimization landscape in high-dimensional settings may be substantially more complex than that induced by carefully designed task objectives.

To scale empowerment to higher-dimensional systems, further methodological advances are required. More robust empowerment estimators, potentially diffusion-based approaches, that do not rely on linear-Gaussian assumptions could enable richer and more expressive probing distributions. Alternatively, exploring fundamentally different formulations of intrinsic motivation may prove fruitful. For example, recent work has proposed an intrinsic metrics based on Kolmogorov–Sinai entropy \cite{shah2026controllableinformationproduction}, a dynamical system level quantity that avoids the ambiguity associated with maintaining separate action distributions. Such directions represent promising avenues for improving scalability, robustness, and conceptual clarity of task-agnostic control frameworks.

\addtolength{\textheight}{-12cm}   














\section*{Acknowledgment}
T.S. was supported by the Koh Family Scholarship. S.T. was supported in part by NSF Award (2513350) and Alliance Innovation Lab in Silicon Valley.

\bibliographystyle{IEEEtranBST/IEEEtran}
\bibliography{refs}

\end{document}